\documentclass[10pt,twocolumn,letterpaper]{article}

\usepackage{cvpr}              % To produce the CAMERA-READY version
\usepackage{graphicx}
\usepackage{amsmath}
\usepackage{amssymb}
\usepackage{booktabs}
\usepackage{bm}
\usepackage{makecell}
\usepackage{utfsym}

\usepackage[accsupp]{axessibility}  % Improves PDF readability for those with disabilities.

\usepackage{enumitem}
\setlist[itemize]{noitemsep, nolistsep}

\usepackage[pagebackref=true,breaklinks=true,colorlinks]{hyperref}

\usepackage[capitalize]{cleveref}
\crefname{section}{Sec.}{Secs.}
\Crefname{section}{Section}{Sections}
\Crefname{table}{Table}{Tables}
\crefname{table}{Tab.}{Tabs.}

\def\cvprPaperID{2289} % *** Enter the CVPR Paper ID here
\def\confName{CVPR}
\def\confYear{2023}

\begin{document}

%%%%%%%%% TITLE - PLEASE UPDATE
\title{Infrastructure Crack Segmentation: 

Boundary Guidance Method and Benchmark Dataset}

% \author{First Author\\
% Institution1\\
% Institution1 address\\
% {\tt\small firstauthor@i1.org}
% % For a paper whose authors are all at the same institution,
% % omit the following lines up until the closing ``}''.
% % Additional authors and addresses can be added with ``\and'',
% % just like the second author.
% % To save space, use either the email address or home page, not both
% \and
% Second Author\\
% Institution2\\
% First line of institution2 address\\
% {\tt\small secondauthor@i2.org}
% }

\author{
Zhili He$^{1,2}$ \; Wang Chen$^{1}$ \; Jian Zhang$^{1}$\footnotemark[1] \; Yu-Hsing Wang$^{2}$\footnotemark[1]\\
$^1$~School of Civil Engineering, Southeast University \\
$^2$~Department of Civil and Environmental Engineering, The Hong Kong University of Science and Technology \\
% {\tt\small guojintao@smail.nju.edu.cn, wangna@smail.nju.edu.cn, qilei@seu.edu.cn, syh@nju.edu.cn}
{\tt\small zl.he@connect.ust.hk, chenwang\_@seu.edu.cn, jian@seu.edu.cn, ceyhwang@ust.hk}
}

% \author{
% Jintao~Guo$^{1}$ \;\; Na Wang$^{1}$ \;\; Lei Qi$^2$\footnotemark[1] \;\; Yinghuan Shi$^{1}$\footnotemark[1] \hspace{0.005cm} \footnotemark[2]\\
% $^1$Nanjing University \hspace{1cm} $^2$Southeast University \\
% {\tt\small \{guojintao, wangna\}@smail.nju.edu.cn, qilei@seu.edu.cn, syh@nju.edu.cn}
% }

\maketitle

\renewcommand{\thefootnote}{\fnsymbol{footnote}}
\footnotetext[1]{Corresponding authors: Jian Zhang and Yu-Hsing Wang.

Preprint submitted to Elsevier.
}
% \footnotetext[2]{
% Jintao Guo, Na Wang and Yinghuan Shi are with the State Key Laboratory for Novel Software Technology and National Institute of Healthcare Data Science, Nanjing University, China.
% Lei Qi is with the School of Computer Science and Engineering, Southeast University, China. 
% }

%%%%%%%%% ABSTRACT
\begin{abstract}
   Cracks provide an essential indicator of infrastructure performance degradation, and achieving high-precision pixel-level crack segmentation is an issue of concern. Unlike the common research paradigms that adopt novel artificial intelligence (AI) methods directly, this paper examines the inherent characteristics of cracks so as to introduce boundary features into crack identification and then builds a boundary guidance crack segmentation model (BGCrack) with targeted structures and modules, including a high frequency module, global information modeling module, joint optimization module, etc. Extensive experimental results verify the feasibility of the proposed designs and the effectiveness of the edge information in improving segmentation results. In addition, considering that notable open-source datasets mainly consist of asphalt pavement cracks because of ease of access, there is no standard and widely recognized dataset yet for steel structures, one of the primary structural forms in civil infrastructure. This paper provides a steel crack dataset that establishes a unified and fair benchmark for the identification of steel cracks.
  BGCrack is available at \textcolor{magenta}{\href{https://github.com/hzlbbfrog/BGCrack}{https://github.com/hzlbbfrog/BGCrack}}.
  
\end{abstract}

\vspace{-0.3cm}
%%%%%%%%% BODY TEXT
\section{Introduction}
\label{sec:intro}
% \vspace{-0.1cm}

Defects inevitably occur, accumulate, and expand during the service life of civil infrastructure due to the complex work environment, thereby posing safety risks to structures and threatening their service life \cite{HE2022104017}. Cracks are among the principal manifestations of multiple structural damage and are characterized by their length, width, and density, and are essential indicators of structural performance degradation as well \cite{DENG2022129238}. Consequently, cracks pose significant safety hazards to infrastructure that cannot be overlooked.

Regular infrastructure inspection is a fundamental approach to detecting damage and evaluating security and service conditions, among which the crack inspection is a crucial item \cite{https://doi.org/10.1111/mice.12421}. However, manual onsite visual detection still occupies the exclusive position in crack inspection, requiring qualified engineers to mark and measure the characteristics of cracks manually \cite{LI201483}. Such a working pipeline has several issues, including high-security risk, high quantitative uncertainty due to human subjectivity, excessive labor cost, exorbitant time overhead, etc. Recently, researchers and practitioners have shown extensive interest in rapidly evolving AI and computer vision (CV) technologies \cite{ZHOU2023104678} and have introduced these technologies into the field of structural defect identification to try to alleviate the aforementioned issues \cite{https://doi.org/10.1111/mice.12263,CHOW2020101105,CHOW2020103372}. These novel technologies also offer the potential to achieve automated and intelligent inspection, which would thoroughly address the challenges associated with manual inspection.

The prevailing research paradigm for AI and CV-based crack inspection involves training deep learning models on crack image datasets and then deploying these models to make inferences from images outside training datasets. The core of this pipeline consists of two elements: deep learning models and crack datasets, both of which are covered in the contributions of this study.

\begin{figure}[tb!]
    \centering
      \includegraphics[width=0.95\linewidth]{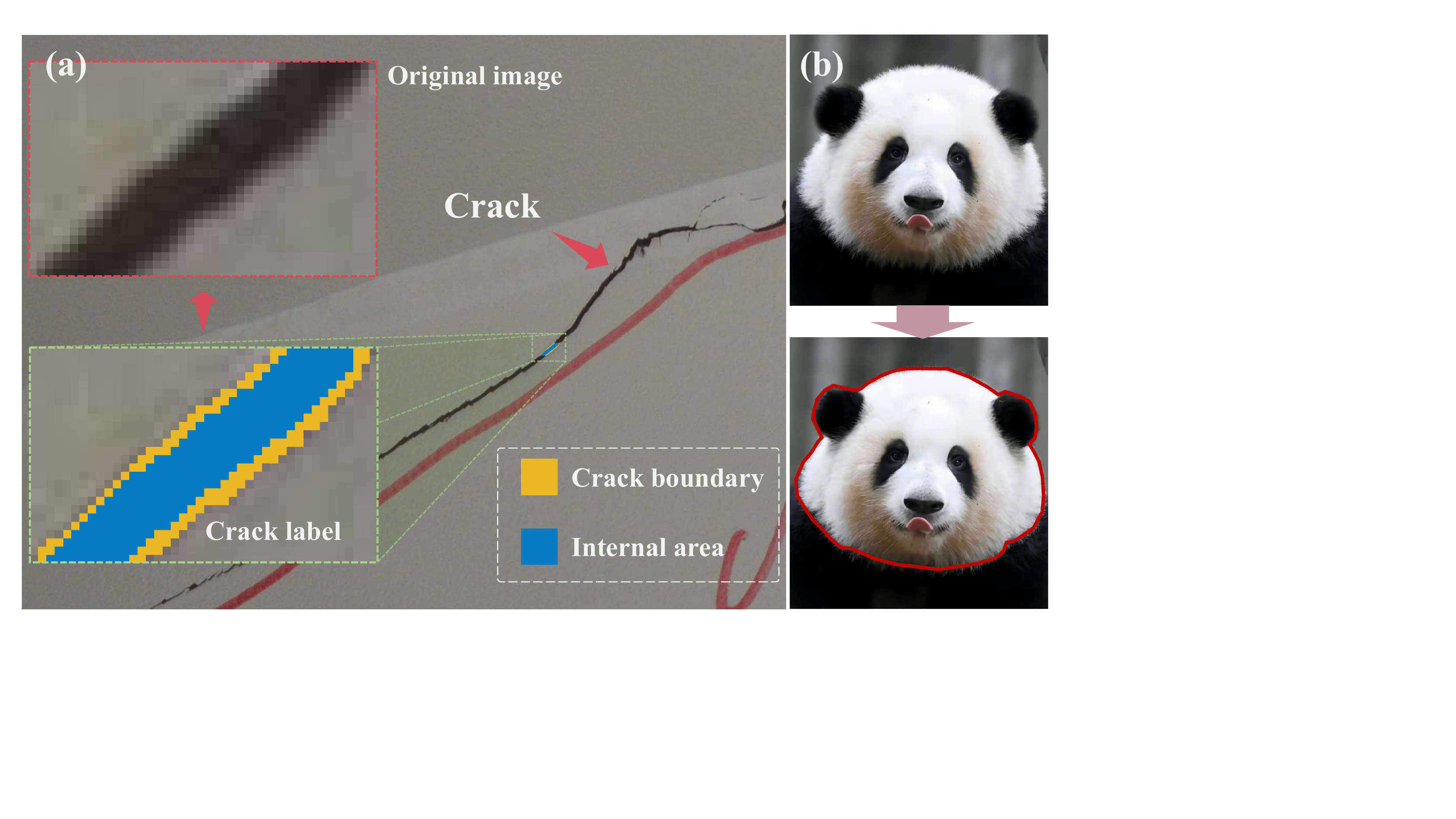}
      \vspace{-0.25cm}
      \caption{Decomposition of images. (a) Cracks can be decomposed into the crack edge and internal area. (b) Edge features help us distinguish and locate the Panda's head.
      }
      \label{fig:one}
    \vspace{-0.3cm}
    \end{figure}

Further, researchers have conducted extensive research to develop deep-learning models that can segment cracks accurately. From the perspective of network structure, these studies can generally be divided into three categories: CNN-based methods \cite{8863123,NI2022101575}, pure Transformer-based methods \cite{GUO2023104646}, and CNN-Transformer hybrid methods \cite{ZHANG2023104712}. These methods pay more attention to the direct application of cutting-edge deep learning methods but neglect the importance of designing targeted methods based on crack intrinsic characteristics. Obviously, in most cases, cracks can be decoupled into two parts: the crack boundary and the internal area (see Fig. \ref{fig:one} (a)). In the human vision system, as shown in Fig. \ref{fig:one} (a) and (b), edge information is always adopted inadvertently to help us distinguish, identify and locate objects \cite{10.1007/978-3-030-58520-4_26}. Some researchers have also leveraged edge features in the salient object detection field and substantiated that integrating boundary features can improve models’ perception accuracy and generate more accurate segmentation images \cite{SCRN,8953756,9008371}. On the other hand, fusing edge information is helpful in getting clear and accurate crack shapes, and Yang et al. \cite{https://doi.org/10.1111/mice.12412} and Dang et al. \cite{DANG2022104472} have demonstrated that it is directly related to the quantitative representation of cracks. Inspired by these findings, this paper designs the BGCrack, which draws boundary information into the field of crack identification and explores and models the relationships between cracks and their edges in an explicit manner.

Accordingly, the main contributions of this study are summarized as follows:
\begin{itemize}[itemsep=2pt,topsep=2pt]
    \item We lead crack boundary features into crack detection and model the relationships between the crack body and crack edge. A novel boundary guidance crack segmentation model named BGCrack is designed.
    \item We create and make available a steel crack dataset and establish a unified and fair benchmark. By doing so, we want to promote standardization and normalization in the field of crack inspection.
\end{itemize}

%-------------------------------------------------------------------------
% \vspace{-0.1cm}
\section{Related Works}
% \vspace{-0.1cm}
\label{sec:related works}
\textbf{Deep learning in the crack inspection.} 
Deep learning models can be classified into three categories based on differences in basic computer vision tasks, namely, image classification, object detection, and semantic segmentation. As for the image classification problem, the objective is to classify defects in given pictures. For instance, Xu et al. \cite{doi:10.1177/1475921720921135} proposed an attribute-based structural damage identification framework based on meta-learning, and Guo et al. \cite{https://doi.org/10.1111/mice.12632} developed a convolutional neural network (CNN) classifier using semi-supervised learning. In terms of the object detection task, the intention is to obtain box-level detection results \cite{https://doi.org/10.1111/mice.12334}. For example, Xu et al. \cite{https://doi.org/10.1002/stc.2313} trained a Faster R-CNN model to identify seismic damage, and Yu et al. \cite{YU2021103514} established a YOLOv4-FPM network to achieve real-time crack detection. Semantic segmentation aims to achieve pixel-level feature representations from input images. Obviously, this approach can determine not only the presence of cracks in an image but also the location and properties of cracks and requires less post-processing than classification. Moreover, for object detection tasks, the highly complex characteristics of cracks make it challenging to determine if predicted bounding boxes are appropriate and accurate for locating cracks. There is almost no unified standard for this assessment. Therefore, semantic segmentation is the most appropriate framework for representing cracks among the three basic computer vision tasks and is also the research focus of this paper.

\textbf{Crack dataset.} 
In addition to conducting research on deep learning models, scholars have also opened sources of some crack image datasets to promote the development of the field. This is because the performance of deep learning models is highly dependent on data as a data-driven technology. Some notable datasets include CrackTree206 \cite{ZOU2012227}, DeepCrackA \cite{LIU2019139}, DeepCrackB \cite{8517148}, CFD \cite{7471507}, and CRACK500 \cite{8694955}. Unfortunately, almost all the images in these datasets are of asphalt pavement cracks because of ease of access. Steel structures are one of the most common structural forms in civil infrastructure. However, there are few influential open-source datasets covering steel cracks. Although many papers have researched the pixel-wise segmentation problem of these cracks, they prefer to leverage their own private datasets because of some engineering disputes or other reasons. To promote standardization and normalization in this field, this paper presents an open-source steel crack dataset. The aim is to establish a unified and impartial benchmark for the identification of steel cracks. The datasets would be available at \textcolor{magenta}{\href{https://github.com/hzlbbfrog/Civil-dataset}{https://github.com/hzlbbfrog/Civil-dataset}}.

% \vspace{-0.1cm}
%-------------------------------------------------------------------------
\section{Methodology}
\label{sec: Method}
\subsection{Motivation and Overall Network Architecture}
\begin{figure*}[tb!]
  \centering
  \includegraphics[width=0.95\linewidth]{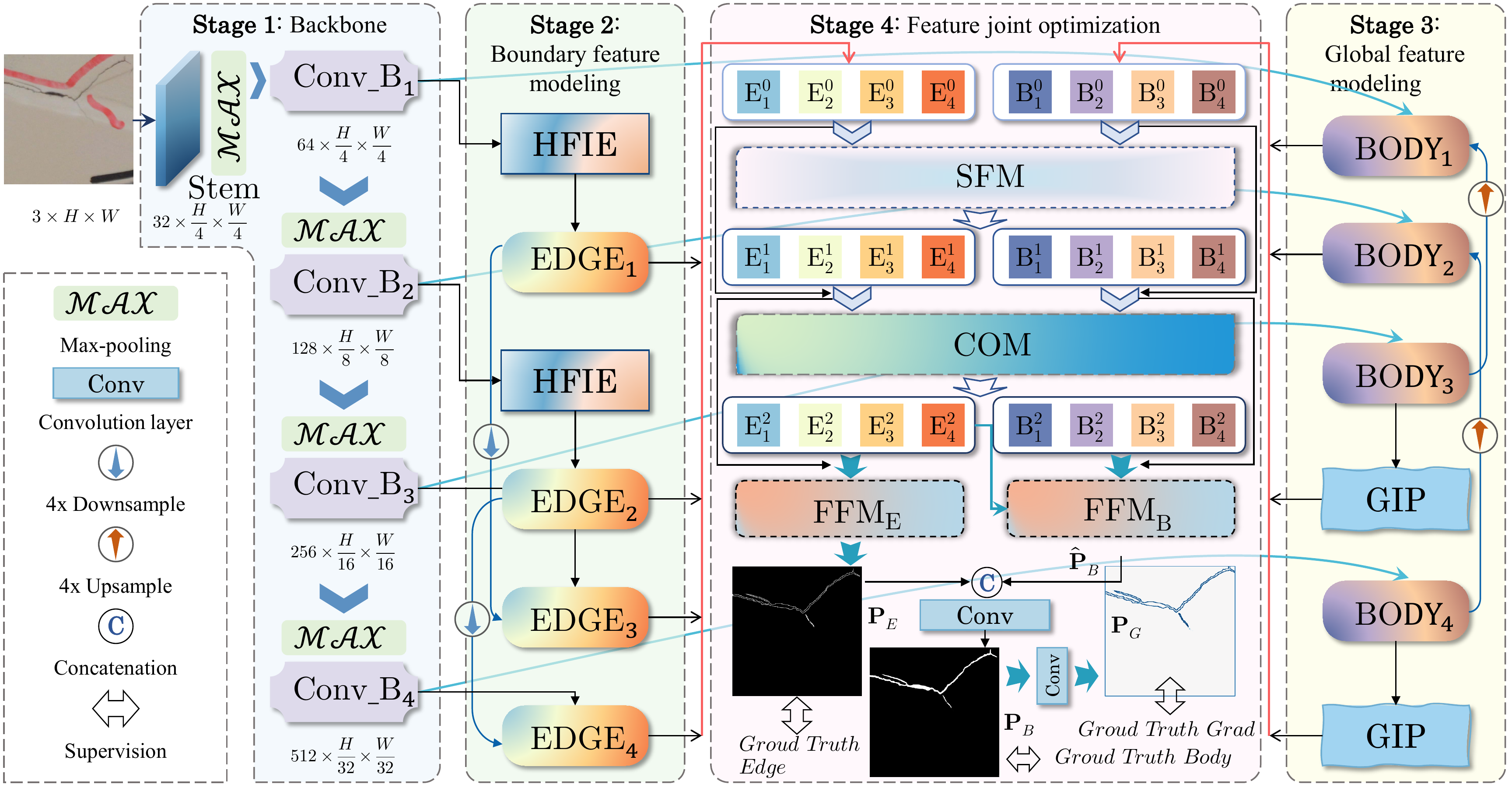}
  \vspace{-0.15cm}
  \caption{
  An overview of the proposed framework.
  }
  \label{fig:two}
  \vspace{-0.25cm}
\end{figure*}

For an input image $\mathbf{I}$ with cracks, the final objective for almost all the crack identification models is to obtain the global features of cracks $\mathbf{I}_{body}$ from $\mathbf{I}$. Based on previous analysis, different from other models, we want BGCrack to learn the representation ability of crack boundary $\mathbf{I}_{edge}$ and $\mathbf{I}_{body}$, simultaneously and further to be able to make $\mathbf{I}_{edge}$ optimize $\mathbf{I}_{body}$. Intuitively, for this purpose, BGCrack is designed in four stages: Stage 1 is the backbone, which is adopted to extract and abstract features, Stage 2 is boundary feature modeling, Stage 3 is global feature modeling, and Stage 4 is the feature optimization. Based on the characteristics of features and purposes of stages, 4 stages are contrapuntally designed in different structures, and the overall framework of BGCrack is illustrated in Fig. \ref{fig:two}. The details are expounded in the following content.

\subsection{Stage 1: Backbone}
The function of the backbone network is to abstract and extract features. The backbone includes two parts, the stem cell, and the hierarchical processing part. The stem part consists of a 3 × 3 convolution with stride 2 and a depth-wise convolution followed by a max pooling operation with a 2 × 2 kernel size and stride of 2. The input image is downsampled 4× to reduce the redundancy inherent in natural images, total parameters, and computation overhead by the stem cell. The hierarchical processing part is composed of 4 stack convolution blocks $(\rm{Conv\_B})$, and each of them includes a convolution layer with the kernel size of 3 × 3 and stride 1 and a depth-wise convolution. A simple non-parametric operator, max pooling, is plugged between two blocks to downsample feature maps similar to U-Net \cite{UNET}. ConvNeXt \cite{CONVNEXT}, the latest research of CNN, empirically demonstrates that (1) fewer activation functions and normalization layers can improve network performance to a certain extent, and (2) large kernel size such as 7 × 7 is more effective than the most widely used size of 3 × 3. Inspired by ConvNeXt, the normalization (Norm) layers and activation functions are re-arranged. Taking the two stack convolution layers as an example, they are structured to “Conv→Norm→Conv→Non-linearity” rather than the most classic structure, “Conv→Norm→Non-linearity→Conv→Norm→Non-linearity”. It is important to note that the nonlinear activation function adopted in BGCrack is SiLU, which is a smoothing function and has more significant gradients than Sigmoid. The batch normalization (BN) is utilized as the normalization layer unless noted specified. Besides, the kernel size of depth-wise convolution layers is all set to 7 × 7. Finally, four-level features can be extracted by Stage 1, \ie, ${\rm{Conv\_B}}_k,\ k=1,2,3,4$.
 
\subsection{Stage 2: Boundary Feature Modeling}
The purpose of Stage 2 is to strip boundary features $\mathbf{I}_{edge}$ from the features modeled by Stage 1. Based on the spectral bias \cite{pmlr-v97-rahaman19a} of deep neural networks (DNNs), we know that DNNs have a learning preference for low-frequency information, and they learn low frequencies faster than high-frequency information. Because the crack edge features are typical high-frequency features, they may not be fully modeled by DNNs, and it is inevitable to lose key information during the training process. A high-frequency information enhancement (HFIE) module is designed to alleviate this problem and better extract high-frequency edge features. Further, because low-level features of CNNs contain more fine-grained detail information represented by edges \cite{9578887}, the HFIE module is just embedded into the low-level features output by Stage 1, \ie, ${\rm{Conv\_B}}_1$ and ${\rm{Conv\_B}}_2$, considering the efficiency of the algorithm. After features pass through HFIE, they are fed into the edge (EDGE) module to conduct the feature embedding to get 2 low-level features, ${\rm EDGE}_1$ and ${\rm EDGE}_2$. For the high-level features, they are sent to the EDGE module, and 2 high-level features can be obtained correspondingly, \ie, ${\rm EDGE}_3$ and ${\rm EDGE}_4$.

\begin{figure*}[tb!]
  \centering
  \includegraphics[width=0.95\linewidth]{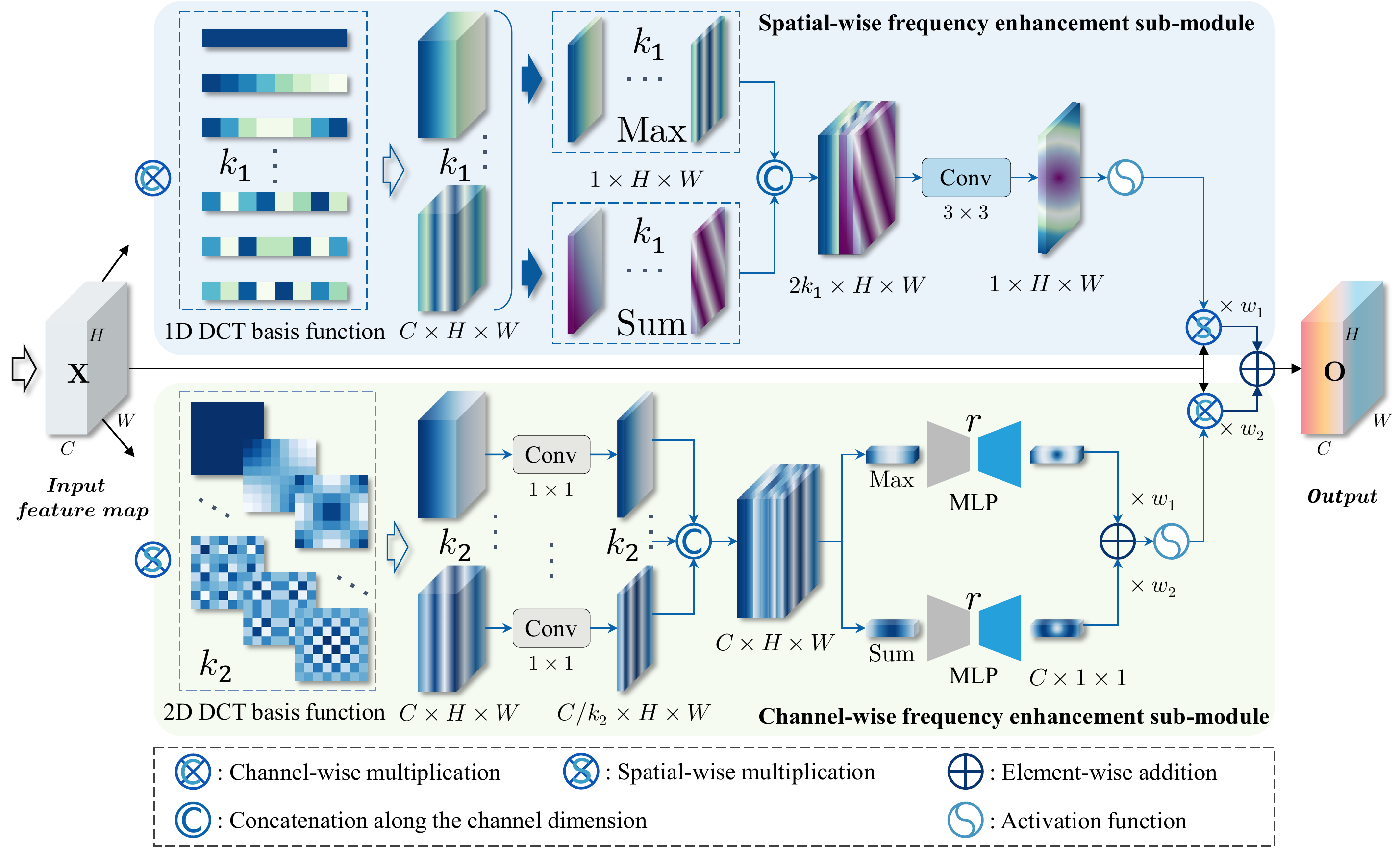}
  \vspace{-0.15cm}
  \caption{
  An overview of the proposed framework.
  }
  \label{fig:three}
  \vspace{-0.25cm}
\end{figure*}

\textbf{HFIE Module.}
The design idea of the HFIE module comes from the FcaNet \cite{FcaNet}, a frequency channel attention network, which firstly proves that attention modules represented by SENet \cite{8701503} usually discard potentially helpful information of high-frequency components except the lowest one because of the global average pooling, a common choice of the deep learning community to compress channel information. Then the authors propose the multi-spectral framework, FcaNet, to extract multi-frequency information to boost network performance using the discrete cosine transform (DCT). However, we find the FcaNet has two drawbacks: (1) It divides the channels of feature maps into different blocks first, then allocates different frequency components to different blocks, which draws into unreasonable prior information; that is, different channels correspond to different frequencies. (2) It just considers channel-wise frequency and ignores spatial-wise frequency. Many research papers have demonstrated that a combination of two-dimensional (2D) information is superior to considering only channel-wise information \cite{CBAMCBAM}. Therefore, the HFIE module is designed to mitigate the aforementioned issues to enhance the representation power of networks, which is illustrated in Fig. \ref{fig:three}.

Precisely, the HFIE module consists of two parallel sub-modules. The first is the spatial-wise frequency enhancement sub-module to convey the transformation, $\mathcal{F}_1:\mathbf{X}\rightarrow\widetilde{\mathbf{O}}\in\mathbb{R}^{C\times H\times W}$. The other one is the channel-wise frequency enhancement sub-module to convey the transformation, $\mathcal{F}_2:\mathbf{X}\rightarrow\hat{\mathbf{O}}\in\mathbb{R}^{C\times H\times W}$. Here $\mathbf{X}\in\mathbb{R}^{C\times H\times W}$ represents the input feature map, and $C$, $H$, and $W$ denote the channel, height, and width, respectively. Combining the $\widetilde{\mathbf{O}}$ and $\hat{\mathbf{O}}$, the output of HFIE $\mathbf{O}$ can be obtained:
\begin{equation}
\mathbf{O}=w_1 \widetilde{\mathbf{O}} \oplus w_2 \widehat{\mathbf{O}}=w_1 \mathcal{F}_1(\mathbf{X}) \oplus w_2 \mathcal{F}_2(\mathbf{X}),
\end{equation}
where $\oplus$ denotes the element-wise addition, $w_1$ and $w_2$ are the learnable parameters.

The spatial-wise frequency enhancement sub-module will enhance frequency related to the spatial dimension. First, $k_1$ one-dimensional (1D) frequency components $\left[f_i\right],i\in\left\{1,2,\ldots,k_1\right\}$ are picked. Then, $k_1$ 1D DCT basis functions $\left[\mathbf{a}^{f_i}\right],i\in\left\{1,2,\ldots,k_1\right\}$ can be computed, where $\mathbf{a}^{f_i}\in\mathbb{R}^C$ means the basis function vector of frequency $f_i$. Next, $k_1$ feature maps with different frequencies can be computed:
\begin{equation}
\mathbf{X}_i^S=\mathbf{X} \odot \mathbf{a}^{f_i}, i \in\left\{1,2, \ldots, k_1\right\},
\end{equation}
where $\mathbf{X}_i^S\in\mathbb{R}^{C\times H\times W}$ is the feature tensor with the basis frequency $f_i$ and $\odot$ represents the channel-wise multiplication. Afterward, a channel-wise max pooling operation $\mathcal{F}_{CMP}$, and a channel-wise sum $\mathcal{F}_{CS}$ are conducted to exact the frequency relationship between channels. Next, a spatial frequency adjustment tensor $\mathbf{O}_{SFA}\in\mathbb{R}^{1\times H\times W}$ can be obtained by concatenating these $2\times k_1$ plane tensor, conducting information interaction by a 3 × 3 convolution layer, and mapping values to (0,1) using a Sigmoid function $\sigma$. Finally, $\widetilde{\mathbf{O}}$ can be computed by the following formula:
\begin{equation}
\mathcal{F}_1(\mathbf{X})=\widetilde{\mathbf{O}}=\mathbf{O}_{S F A} \circledast \mathbf{X},
\end{equation}
here $\circledast$ represents the spatial-wise multiplication.

The channel-wise frequency enhancement sub-module wants to enhance frequency related to the channel dimension. Firstly, $k_2$ 2D frequency components $\left[\left(f_i^H,f_i^W\right)\right],i\in\left\{1,2,\ldots,k_2\right\}$ are selected. Based on these, $k_2$ 2D DCT basis functions $\left[\mathbf{A}^{\left(f_i^H,f_i^W\right)}\right],i\in\left\{1,2,\ldots,k_2\right\}$ can be gotten, here $\mathbf{A}^{\left(f_i^H,f_i^W\right)}\in\mathbb{R}^{H\times W}$ denotes the basis function plane array of spatial frequency $\left(f_i^H,f_i^W\right)$. Next, $k_2$ feature maps with different frequencies can be computed:
\begin{equation}
\mathbf{X}_i^C=\mathbf{X} \circledast \mathbf{A}^{\left(f_i^H, f_i^W\right)}, i \in\left\{1,2, \ldots, k_2\right\},
\end{equation}
where $\mathbf{X}_i^C\in\mathbb{R}^{C\times H\times W}$ is the feature tensor. Then $k_2$ point-wise convolution layers are applied to fuse channel contextual information and compress the number of channels to ${C / k_2}$:
\begin{equation}
\mathbf{X}_i^{C / k_2}=\mathrm{PWConv}_i \left(\mathbf{X}_i^C\right), i \in\left\{1,2, \ldots, k_2\right\},
\end{equation}
here $\mathbf{X}_i^{C / k_2}\in\mathbb{R}^{\left(C / k_2\right)\times H\times W}$ and $\mathrm{PW\ {Conv}}_i$ is $i$-th point-wise convolution. Afterwards, concatenating $k_2$ $\mathbf{X}_i^{C / k_2}$, the number of channels can be recovered, and $\mathbf{X}^C\in\mathbb{R}^{C\times H\times W}$ with different frequency components can be obtained. Then, a global max pooling operation $\mathcal{F}_{GMP}$, and a space-wise sum operation $\mathcal{F}_{SS}$ are conducted to exact the frequency relationship along spatial directions, and 2 vectors $\bm{max}\in\mathbb{R}^{C\times1\times1}$ and $\bm{sum}\in\mathbb{R}^{C\times1\times1}$ can be achieved. 

Later, two multi-layer perceptrons (MLPs) are plugged as channel context semantic aggregators to achieve point-wise-channel interactions with a reduction ratio $r$, which is set to 8 like  \cite{HE2022104017}. Similar to the previous sub-module, a Sigmoid function is adopted to map the values to $\left(0,1\right)$. Further, two trainable parameters $w_1$ and $w_2$ are added to adjust the weights adaptively:
\begin{equation}
\mathbf{O}_{C F A}=\sigma\left[w_1 \times\left(\mathrm{MLP}_1(\boldsymbol{m a x})\right)+w_2 \times\left(\mathrm{MLP}_2(\boldsymbol{s u m})\right)\right].
\end{equation}
Eventually, the output $\hat{\mathbf{O}}$ can be computed by the following equation:
\begin{equation}
\mathcal{F}_2(\mathbf{X})=\widehat{\mathbf{O}}=\mathbf{O}_{C F A} \odot \mathbf{X}.
\end{equation}
It is important to note that $k_1$ and $k_2$ are set to 8, 1D frequency components are set to $\left[0\times\left\lfloor\frac{C}{8}\right\rfloor,\ldots,7\times\left\lfloor\frac{C}{8}\right\rfloor\right]$, and 2D frequency components are set to$\left[\left(0\times\left\lfloor\frac{H}{8}\right\rfloor,0\times\left\lfloor\frac{W}{8}\right\rfloor\right),\ldots,\left(7\times\left\lfloor\frac{H}{8}\right\rfloor,7\times\left\lfloor\frac{W}{8}\right\rfloor\right)\right]$ simply in the all the experiments. Obviously, the choice of the fundamental frequency is not unique. FcaNet has conducted a very complex parameter selection to achieve the best results for the specific dataset based on the selected frequency set, but the frequency set is not universally representative. HFIE takes the most simple and fundamental set to show the generalization and effectiveness of the algorithm. It is obvious that the performance can be improved further after parameter optimization like FcaNet.

\subsection{Stage 3: Global Feature Modeling}
The purpose of Stage 3 is to model global semantic features $\mathbf{I}_{body}$ from the features extracted by Stage 1. Similar to Stage 2, 4 modules named the body module (BODY) are designed to embed features from Stage 1, and the corresponding multi-level features are ${\rm BODY}_k,\ k=1,2,3,4$. For design simplicity and consistency, the architecture of the BODY module is the same as that of the EDGE module.

\begin{figure*}[tb!]
  \centering
  \includegraphics[width=0.95\linewidth]{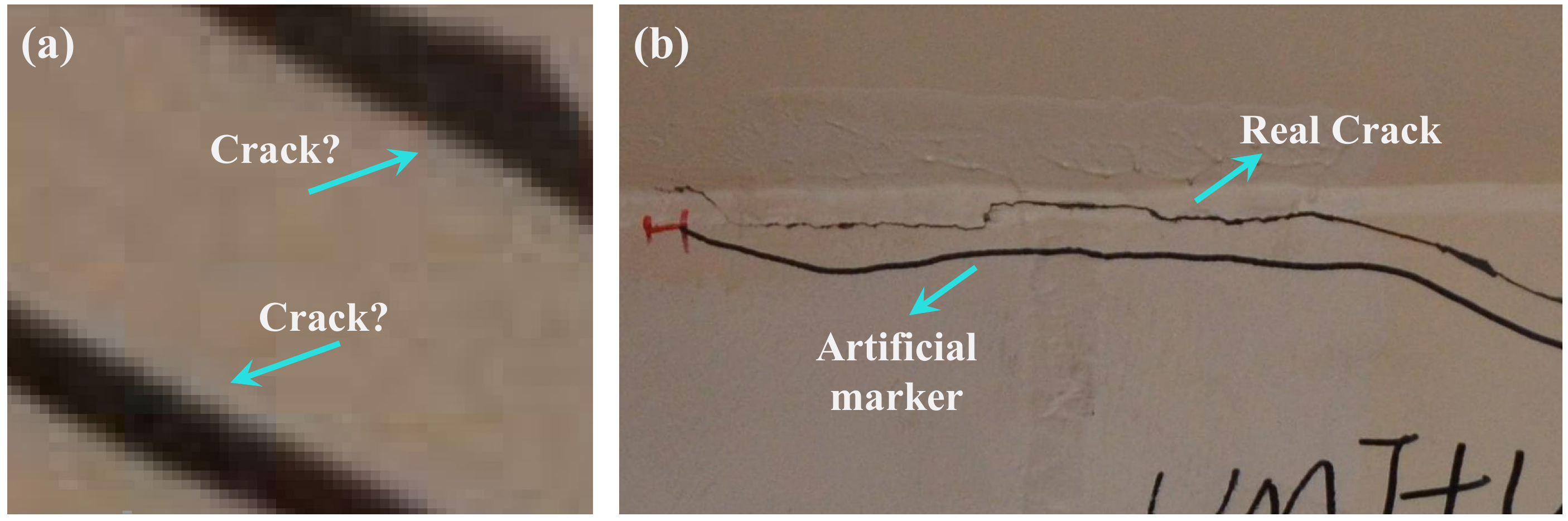}
  \vspace{-0.15cm}
  \caption{
  Comparison between local information and long-range semantic information. (a) Local information. (b) Long-range semantic information.
  }
  \label{fig:four}
  \vspace{-0.25cm}
\end{figure*}

However, unlike Stage 2, the global feature modeling of cracks needs to pay attention to the long-term context information, which can be roughly explained by the recognition process of the human visual system. As shown in Fig. \ref{fig:four} (a), it is difficult to judge which one is the real crack just based on the short-range features. However, we can quickly know that the top one is the crack and the below one is an artificial marker once taking the more context semantic information and the long-term dependency relationship of pixels into consideration (see Fig. \ref{fig:four} (b)), namely, we can model the global features of cracks more accurately. To achieve this, a novel global information perception (GIP) module is designed to capture and model long-distance semantic information better and enhance the perception capability of networks for global information. Additionally, because high-level features of CNNs have a larger receptive field, which brings more redundant long-range context information \cite{9578887,YU2017235}, embedding GIP into high-level features is more effective. Therefore, ${\rm BODY}_3$ and ${\rm BODY}_4$ are appended by a GIP module, respectively. On the other hand, high-level features have a smaller spatial resolution, so this plugging scheme can also bring less calculation quantity.

\textbf{GIP module.} Motivation. The inherent locality of convolution operations causes CNNs to be experts in modeling local information but have congenital limitations and yield subpar performances in capturing long-term context features \cite{chen2021transunet}, such as cracks’ global semantic information that shows massive changes with respect to modality and size. Therefore, how to improve the global perception capability of networks has become a hotspot issue and has drawn extensive attention from academic and industrial circles. 

An effectual method is ushering the ingredients with global awareness into native CNNs, such as the Fast Fourier Transform (FFT) algorithm, which can model long-range context information from a view of the frequency domain. Some influential models have been proposed based on FFT, represented by LaMa \cite{LAMA}, a compelling image inpainting model using Fast Fourier Convolution (FFC) \cite{FFC}, and DeepRFT \cite{mao2022intriguing}, which is an advanced image deblurring framework, utilizing residual convolution blocks with FFT. Inspired by the success of FFT in CNNs, it is also drawn into the GIP module to model global contexts in the frequency domain.

Additionally, another way is to substitute CNNs with other units, among which the most representative framework is Transformer \cite{vaswani2017attention}, which can capture long-term dependencies because the self-attention mechanism in it can bring intrinsically global perception without the complex hierarchical stacking structures like CNNs. Transformer has achieved tremendous and widespread success in multiple fields, such as natural language processing \cite{vaswani2017attention}, machine translation \cite{devlin-etal-2019-bert}, and computer vision \cite{dosovitskiy2020image}. Specifically, it has been regarded as one of the most significant research progress in the field of computer vision, and some frameworks based on Transformer have also made an enormous impact, such as ViT \cite{dosovitskiy2020image}, DPT \cite{9711226}, and Swin Transformer \cite{9710580}. Therefore, Transformer is also led into the GIP module to learn long-range relations in the pixel domain, aka the time domain.

\begin{figure*}[tb!]
  \centering
  \includegraphics[width=0.95\linewidth]{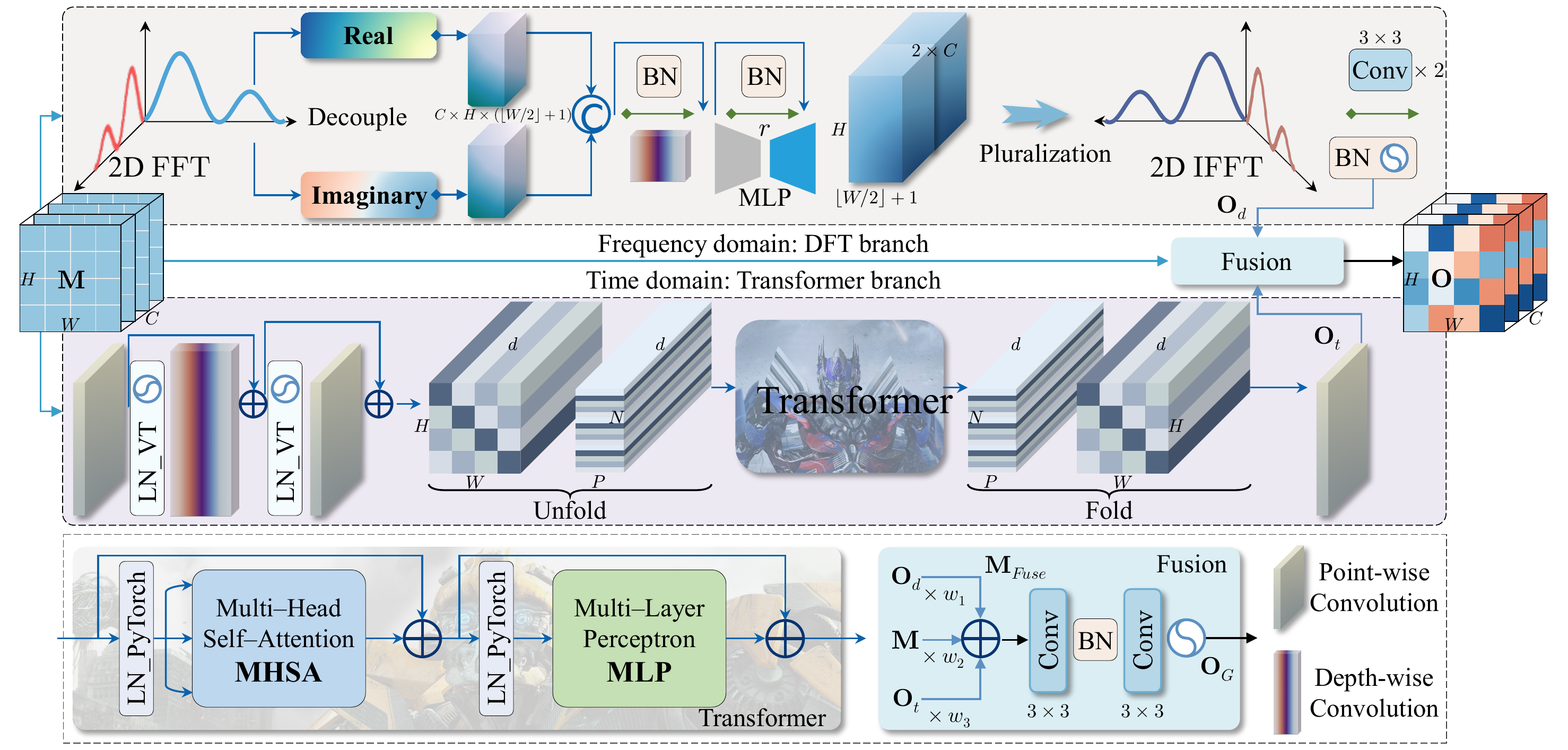}
  \vspace{-0.15cm}
  \caption{
  The architecture of the global information perception (GIP) module. LN\_VT is the layer normalization (LN) method used in most vision transformer frameworks. LN\_PyTorch is the default LN in PyTorch. The difference is that the affine transformation parameters of LN\_VT are channel-wise, and the parameters of LN\_PyTorch are element-wise.
  }
  \label{fig:five}
  \vspace{-0.25cm}
\end{figure*}

Then, the overall structure of the GIP module can be acquired, which is resented in Fig. \ref{fig:five} and consists of two branches: The first one is named the frequency domain branch, in which the GIP module performs Discrete Fourier Transform (DFT) on a feature map by the FFT algorithm to provide the global perception. The second one is the time domain branch, in which Transformer is plugged to extract long-distance context dependency. Then a fusion model is designed to integrate the input and two outputs. To the authors' best knowledge, the GIP module takes the lead in integrating global information from the frequency domain and time domain organically, which can enhance the representation power for global features of cracks effectively to improve identification accuracy. The detailed design is illustrated below.

DFT branch. Let the input feature map be $\mathbf{M}\in\mathbb{R}^{C\times H\times W}$. (1) Performing 2D DFT on $\mathbf{M}$ by the 2D real FFT algorithm to get the complex spectrum:
\begin{equation}
\mathcal{F}(\mathbf{M})=\operatorname{RFFT}(\mathbf{M}),
\end{equation}
where $\mathcal{F}\left(\mathbf{M}\right)\in\mathbb{C}^{C\times H\times\left(\left\lfloor{W/2}\right\rfloor+1\right)}$ is the complex spectrum, here the conjugate symmetry of spectrums is adopted to compress the width dimension, and RFFT\ represents the 2D real FFT algorithm. (2) Decoupling the real and imaginary parts of the complex spectrum to gain $\mathcal{R}\left(\mathcal{F}\left(\mathbf{M}\right)\right)\in\mathbb{R}^{C\times H\times\left(\left\lfloor{W/2}\right\rfloor+1\right)}$, and $\mathcal{I}\left(\mathcal{F}\left(\mathbf{M}\right)\right)\in\mathbb{R}^{C\times H\times\left(\left\lfloor{W/2}\right\rfloor+1\right)}$, respectively. (3) Concatenating $\mathcal{R}\left(\mathcal{F}\left(\mathbf{M}\right)\right)$ and $\mathcal{I}\left(\mathcal{F}\left(\mathbf{M}\right)\right)$ along the channel direction and using a block to model the new compound features. The design of the block refers to the MetaFormer \cite{METAFORMER}. It can be simply described as “BN→$\rm{DW\ Conv}$→BN→$\rm{PW\ {Conv}}_1$→SiLU→$\rm{PW\ {Conv}}_2$”, where $\rm{DW Conv}$ denotes the depth-wise convolution with the 7 × 7 kernel size, $\rm{PW\ {Conv}}_1$ compresses the number of channels to a quarter of its previous size, and the function of $\rm{PW\ {Conv}}_2$ is to restore the number of channels. Next, transforming the real features into the complex field to get a new complex spectrum $\widetilde{\mathcal{F}}\left(\mathbf{M}\right)\in\mathbb{C}^{C\times H\times\left(\left\lfloor{W/2}\right\rfloor+1\right)}$. (4) Performing inverse 2D real FFT to convert the frequency domain into the time domain. Then, the output feature map $\mathbf{O}_d\in\mathbb{R}^{C\times H\times W}$ can be acquired after a convolution block with 2 3 × 3 convolutions.

Transformer branch. Although Transformer has many advantages compared with CNNs, every coin has two sides. Transformer’s design has an inherent defect that lacks the spatial inductive bias, the intrinsic property of CNNs \cite{xiao2021early}, which causes that: (1) Transformer needs more parameters to lean visual representations and is heavy-wight \cite{mehta2022mobilevit}, {\rm \eg}, ViT-L/16 vs. MobileNetv3 \cite{9008835}: 307M vs. 7.5 M parameters, and DPT-Large vs. DeepLabv3 \cite{chen2017rethinking}: 343M vs. 59 M parameters. (2) It needs to be trained on sufficient amounts of data to generalize well \cite{dosovitskiy2020image}, {\rm \eg}, ViT is trained on the JFT-300M dataset containing over 300M images, and DeiT \cite{pmlr-v139-touvron21a} relies on extensive data augmentation tricks. These drawbacks make Transformer need rich computing resources and unfriendly for tasks without rich data, such as the crack identification task involved in this paper. Therefore, some researchers propose CNN-Transformer hybrid models meriting both the local perception and spatial inductive bias of CNN and the global modeling ability of Transformer to apply to downstream tasks successfully, {\rm \eg}, TransUNet \cite{chen2021transunet}for medical image segmentation and MSHViT \cite{HAURUM2022104614} for defect detection. Essentially, BGCrack is also a hybrid solution, and Transformer is embedded in a CNN. On the other hand, some authors try to modify the Transformer and usher the spatial inductive bias into the vanilla Transformer. A representative method is MobileViT \cite{mehta2022mobilevit}, which implicitly incorporates convolution-like characteristics into Transformer to make it learn global representations with the spatial bias to achieve the goals of lightweight and better generalization ability \cite{mehta2022mobilevit}. Inspired by MobileViT, a similar lightweight Transformer block is exploited to substitute the vanilla Transformer.

As shown in Fig. \ref{fig:five}, let the input feature map be $\mathbf{M}\in\mathbb{R}^{C\times H\times W}$ as well. (1) a point-wise convolution is utilized to project the input features into a high-dimensional space to get a d-dimensional tensor $\mathbf{M}_1\in\mathbb{R}^{d\times H\times W}$, where $d>C$. (2) A convolution block containing two layer normalization (LN) layers, a $n\times n$ depth-wise convolution layer, and a point-wise convolution layer is appended to get the local representation $\mathbf{M}_2\in\mathbb{R}^{d\times H\times W}$. (3) In order to make this module to be able to earn global information with the spatial inductive bias, $\mathbf{M}_2$ is unfolded into N non-overlapping flattened patches in the spatial dimension to acquire $\mathbf{M}_U\in\mathbb{R}^{N\times P\times d}$, where $N=\frac{H\times W}{P}$ represents the number of patches, $P=h_P\times w_P$ denotes the area of patches, and $h_P$ and $w_P$ means the height and width of patches, respectively. The following Transformer would calculate the attention among patches to learn global representations. However, it doesn’t encode the local information in each patch, which induces there are “holes” in the modeled global attention. Therefore, with an eye to extracting full long-range contexts, the local receptive field brought by the depth-wise convolution in (1) needs to cover patches, \ie, $h_P\le n$ and $w_P\le n$. (4) For every vector $\mathbf{v}_p\in\mathbb{R}^{1\times1\times d},p\in\left\{1,2,\ldots,P\right\}$ within each patch, the inter-patch relationships are modeled by Transformer, and $P$ global correlation feature maps $\mathbf{M}_T^p\in\mathbb{R}^{N\times1\times d}$ are gained:
\begin{equation}
\mathbf{M}_T^p=\text { Transformer }\left(\mathbf{M}_U^p\right),
\end{equation}
where $\mathbf{M}_U^p\in\mathbb{R}^{N\times1\times d}$ denotes the combination of all the $p$-th vector $\mathbf{v}_p$ in $N$ patches. Concatenating all the $\mathbf{M}_T^p$ along the patch dimension, the feature map after the Transformer can be acquired, let it be $\mathbf{M}_T\in\mathbb{R}^{N\times P\times d}$. Because the spatial locations of pixels in each patch and inter-patch spatial relationships are retained, it keeps spatial inductive bias compared with the conventional Transformer. (5) Folding $\mathbf{M}_T$ to recover the shape of the feature map before unfolding to achieve $\mathbf{M}_F\in\mathbb{R}^{d\times H\times W}$. (6) A point-wise convolution is adopted to embed $\mathbf{M}_F$ into $C$-dimensional to get the output features $\mathbf{O}_t\in\mathbb{R}^{C\times H\times W}$.

Fusion. A fusion block is designed to fuse an input $\mathbf{M}$ and two outputs $\mathbf{O}_d$ and $\mathbf{O}_t$. First, three learnable parameters $w_1$, $w_2$, and $w_3$ are added to adjust the weights:
\begin{equation}
\mathbf{M}_{\text {Fuse }}=w_1 \times \mathbf{O}_d+w_2 \times \mathbf{M}+w_3 \times \mathbf{O}_t.
\end{equation}
Then a simple convolution block with the structure of “3 × 3 Conv→BN→3 × 3 Conv→SiLU” is applied to model the fused feature map $\mathbf{M}_{Fuse}$ to obtain the final output of the GIP module $\mathbf{O}_G\in\mathbb{R}^{C\times H\times W}$.

\begin{figure*}[tb!]
  \centering
  \includegraphics[width=0.9 \linewidth]{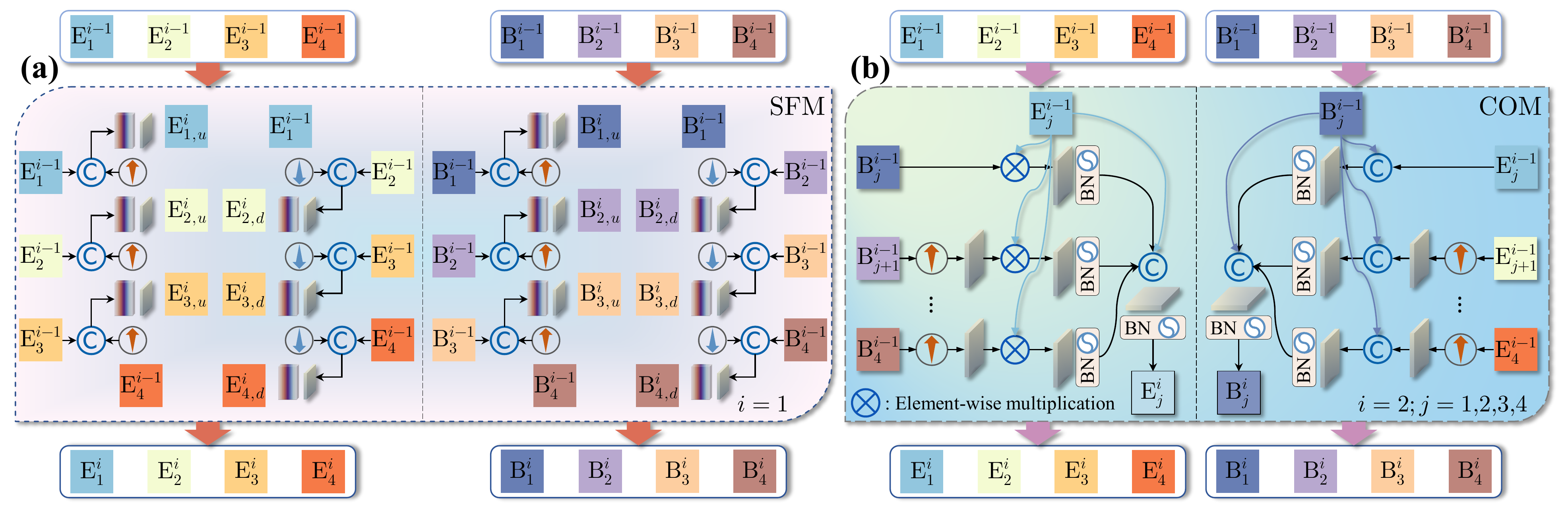}
  \vspace{-0.15cm}
  \caption{
  The structures of the proposed modules. (a) Self-fusion module (SFM). (b) Cross optimization module (COM).
  }
  \label{fig:six}
  \vspace{-0.25cm}
\end{figure*}

\subsection{Stage 4: Feature Joint Optimization}
This stage is the fusion and joint optimization stage of the boundary features $\mathbf{I}_{edge}$ output by Stage 2 and the global features $\mathbf{I}_{body}$ obtained by Stage 3. To simplify notations, let the 8 output features of two stages be $E_j^0,\ j=1,2,3,4$ and $B_j^0,\ j=1,2,3,4$, respectively. Here $E_j^i$ denotes the $j$-th boundary feature map in the $i$-th calculation phase, and $i=0,1,2$ (see Fig. \ref{fig:two} above). For the same reason, we have $B_j^i$. Besides, the dense connection design is incorporated into this stage to improve the information flow, strengthen the feature propagation and maintain network gradients. Each set of tensors is directly added to all the corresponding tensors before and then input to the next module (see Fig. \ref{fig:two}).

In this stage, first of all, a self-fusion module (SFM) is designed to promote information interaction and fusion of multi-level and multi-scale features inside the boundary feature set and global feature set to get $E_j^1,\ j=1,2,3,4$ and $B_j^1,\ j=1,2,3,4$. Afterward, because the boundary features are helpful to improve the body features, and the body features are also beneficial for refining the boundary features, a cross optimization module (COM) is further developed to promote the integration between the two kinds of features to gain $E_j^2,\ j=1,2,3,4$ and $B_j^2,\ j=1,2,3,4$. After that, two feature fusion modules (FFM), \ie, ${\rm FFM}_E$ for edge features, and ${\rm FFM}_B$ for body features, are proposed to align the spatial dimensions, namely, recover the size from inputting size, $H\times W$, and the corresponding outputs $\mathbf{O}_E\in\mathbb{R}^{1\times H\times W}$ and $\mathbf{O}_B\in\mathbb{R}^{1\times H\times W}$ are achieved.

The detailed interpretation of the architectures of the above-mentioned modules, \ie, SFM, COM, and FFM, can be seen in the next subsections.

SFM. The structure of the SFM is presented in Fig. \ref{fig:six} (a). The fusion of the boundary features is the same as that of the global features, so, for simplicity, we introduce the fusion of the boundary features. The fusion includes two branches, namely, an upsampling branch and a downsampling branch. In the upsampling branch, the spatial size of the feature map is gradually enlarged, and the low-level feature map is fused with the current features simultaneously. By this branch, the high-level and large-scale information can spread to low-level and small-scale feature maps, and new tensors are gained, $E_{1,u}^1$, $E_{2,u}^1$, and $E_{3,u}^1$. The purpose and process of the downsampling branch are exactly the opposite of the upsampling branch. In the downsampling branch, the low-level and small-scale information is integrated into the high-level and large-scale feature maps, and correspondingly, $E_{2,d}^1$, $E_{3,d}^1$, and $E_{4,d}^1$ are gotten. The final output of the SFM is the addition of the upsampling features and downsampling features simply: $E_1^1=E_{1,u}^1+E_1^0$, $E_2^1=E_{2,u}^1+E_{2,d}^1$, $E_3^1=E_{3,u}^1+E_{3,d}^1$, and $E_4^1=E_4^0+E_{4,d}^1$.

COM. Before designing the COM, the relationship between $\mathbf{I}_{body}$ and $\mathbf{I}_{edge}$ needs to be clarified in advance. As introduced before, a crack can be decoupled into two parts, namely, the crack boundary $\mathbf{I}_{edge}$ and internal area $\mathbf{I}_{inner}$. Therefore, the logical interrelations of crack features can be obtained:
\begin{equation}
\mathbf{I}_{b o d y}=\mathbf{I}_{b o d y} \cup \mathbf{I}_{\text {edge }}.
\label{body}
\end{equation}
Further, we have
\begin{equation}
\mathbf{I}_{\text {edge }}=\mathbf{I}_{\text {edge }} \cap \mathbf{I}_{\text {body }}.
\label{edge}
\end{equation}
Based on the logical interrelations, it is clear that on the one hand, $\mathbf{I}_{edge}$ can improve the accuracy of $\mathbf{I}_{body}$, and on the other hand, $\mathbf{I}_{body}$ is able to refine $\mathbf{I}_{edge}$ [21], which is also the design motivation of COM. Because Eq. \eqref{edge} are difficult to represent by a CNN, they are further converted to the following equation equivalently:
\begin{equation}
\mathbf{I}_{\text {edge }}=\mathbf{I}_{\text {edge }} \otimes \mathbf{I}_{\text {body }},
\end{equation}
where $\otimes$ denotes element-wise multiplication. Based on the equations, the COM is designed, and the details are shown in Fig. \ref{fig:six} (b).

\begin{figure*}[tb!]
  \centering
  \includegraphics[width=0.95\linewidth]{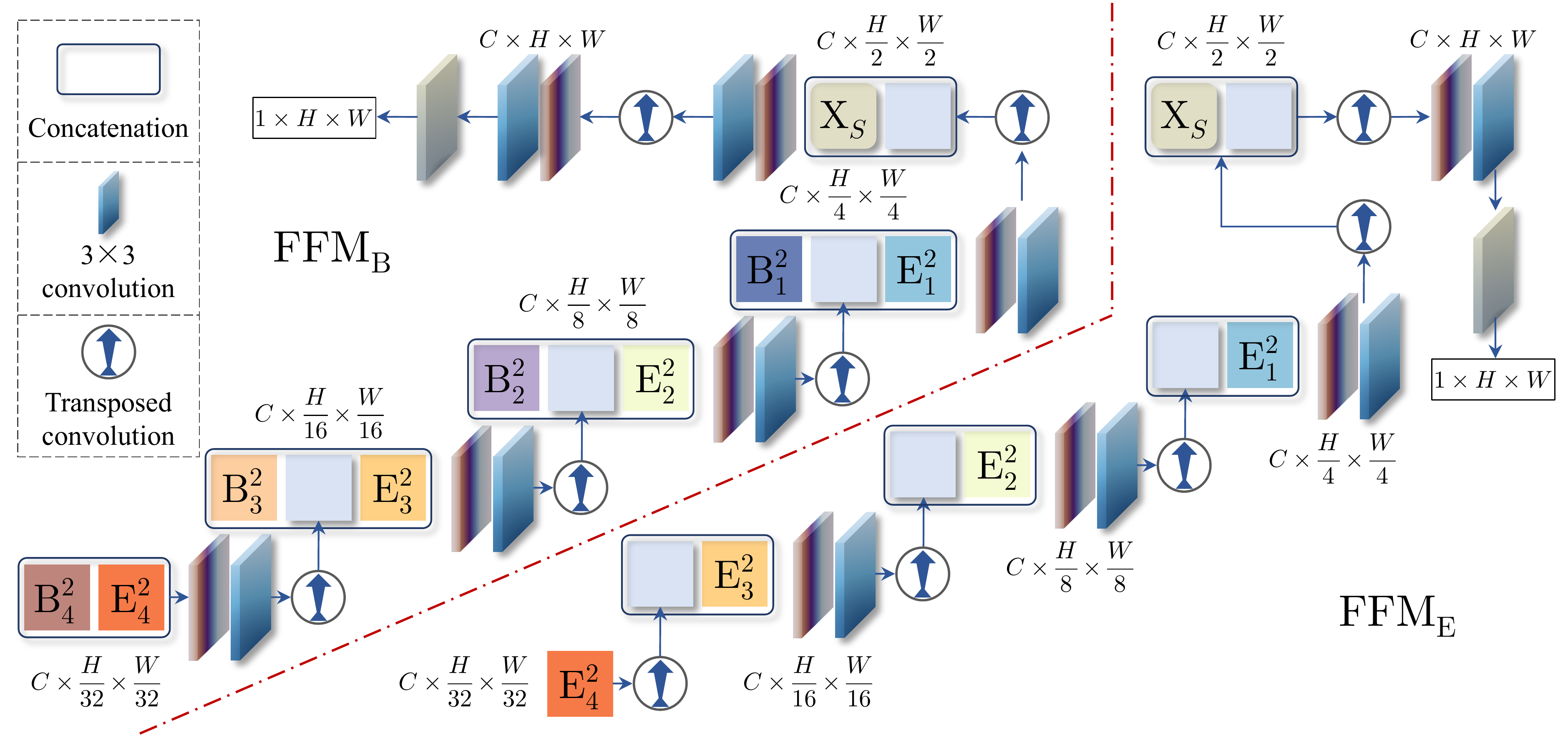}
  \vspace{-0.15cm}
  \caption{
  The structures of the proposed modules. (a) Self-fusion module (SFM). (b) Cross optimization module (COM).
  }
  \label{fig:seven}
  \vspace{-0.25cm}
\end{figure*}

Specifically, the COM can be defined as
\begin{subequations}
\begin{align}
\mathrm{E}_j^2=\mathcal{E}\left(\mathrm{E}_j^1,\left\{\mathrm{~B}_k^1\right\}_{k=j}^4\right),\label{E}\\
\mathrm{B}_j^2=\mathcal{B}\left(\mathrm{B}_j^1,\left\{\mathrm{E}_k^1\right\}_{k=j}^4\right),\label{B}
\end{align}
\end{subequations}
where $\mathcal{E}$ and $\mathcal{B}$ represent the boundary feature optimization function and global feature optimization function, respectively. Because the lower-level features contain more distractors and background information, and the higher-level features can extract semantic and discriminative information with few distractors \cite{SCRN}, the current features are optimized by the same level or higher-level features, {\rm \eg}, $\left\{B_k^1\right\}_{k=j}^4$ are adopted to fine $E_j^2$, instead of the lower-level $\left\{B_k^1\right\}_{k=1}^{j-1}$. Therefore, the Eq. \eqref{E} can be expanded as
\begin{equation}
\mathrm{E}_j^2=w e_0^j \mathrm{E}_j^1+w e_1^j \widehat{\mathrm{E}}_j^1+w e_2^j \widehat{\mathrm{E}}_{j+1}^1+\cdots+w e_{4-j+1}^j \widehat{\mathrm{E}}_4^1,
\end{equation}
where $\mathbf{we}=\left({we}_0^j,{we}_1^j,\ldots,{we}_{4-j+1}^j\right)$ is a learnable weight parameter vector. The Eq. \eqref{B} can also be expanded as
\begin{equation}
\mathrm{B}_j^2=w b_0^j \mathrm{~B}_j^1+w b_1^j \hat{\mathrm{B}}_j^1+w b_2^j \hat{\mathrm{B}}_{j+1}^1+\cdots+w b_{4-j+1}^j \hat{\mathrm{B}}_4^1,
\end{equation}
where $\mathbf{wb}=\left({wb}_0^j,{wb}_1^j,\ldots,{wb}_{4-j+1}^j\right)$ is another learnable weight parameter vector.

FFM. To put it simply, FFM is a hierarchical and multi-level feature fusion module whose designs are inspired by U-Net. For integrating the boundary and global features, two similar blocks, \ie, ${\rm FFM}_E$ and ${\rm FFM}_B$, are developed (see Fig. \ref{fig:seven} above). Next, taking the ${\rm FFM}_B$ as an example, introduce the FFM’s structure. Because global features are the final purpose, ${\rm FFM}_B$ not only fuse 4 global sub-features, $B_j^2,\ j=1,2,3,4$, but also integrate 4 edge sub-features, $E_j^2,\ j=1,2,3,4$, to refine the global output further. Firstly, it starts from the highest-level features $B_4^2$ and $E_4^2$, and concatenates them along the channel dimension. Then, a depth-wise convolution and a standard 3 × 3 convolution are appended to extract information. Next, a transposed convolution is adopted to enlarge the combined feature map to twice its original spatial size. The above process is repeated until the lowest-level features $B_1^2$ and $E_1^2$ are fused. After that, $X_S$, output from the stem part, is integrated further to supplement details. Finally, a point-wise convolution layer is utilized to depress the number of channels to 1. For the convenience of differentiation, we let the outputs be $\mathbf{P}_E\in\mathbb{R}^{1\times H\times W}$ and $\hat{\mathbf{P}}_B\in\mathbb{R}^{1\times H\times W}$, respectively. Further, $\mathbf{P}_E$ is integrated into the $\hat{\mathbf{P}}_B$ to refine the global features to get the final global feature output ${\mathbf{P}}_B$, as shown in Fig. \ref{fig:two}.

\subsection{Loss Function}
First of all, the most classic binary cross-entropy (BCE) loss is applied to the similarity assessment between the prediction and ground-truth. The BCE loss between the global output and the corresponding ground-truth is
\begin{equation}
\begin{aligned}
\mathcal{L}_{\mathrm{BCE}}^B
&=\frac{1}{N} \sum_{i=1}^N \left\{ -\frac{1}{HW} \sum_{(h, w)}\left[\mathbf{G}_B^i(h, w) \times \log \left({\mathbf{P}}_B^i(h, w)\right)\right] \right.\\
&-\frac{1}{HW} \left. \sum_{(h, w)}\left[\left(1-\mathbf{G}_B^i(h, w)\right) \times \log \left(1-{\mathbf{P}}_B^i(h, w)\right)\right] \right\},
\end{aligned}
\end{equation}

\begin{figure*}[tb!]
  \centering
  \includegraphics[width=0.95\linewidth]{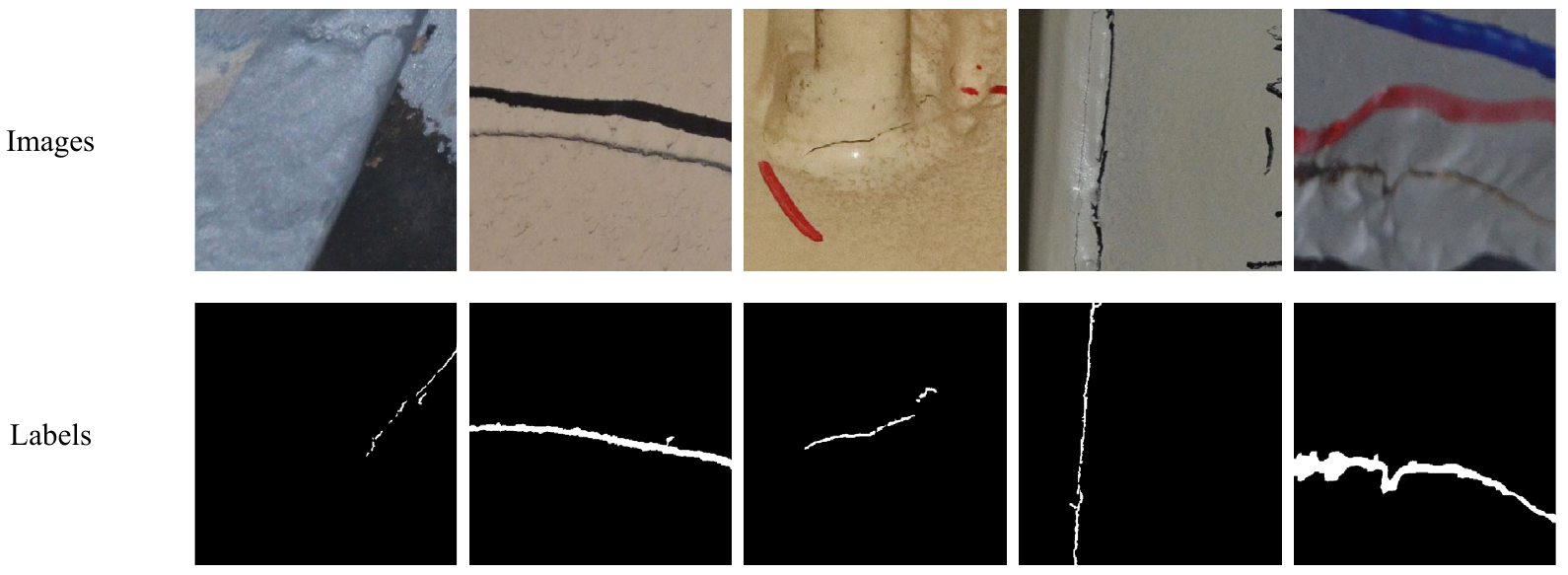}
  \vspace{-0.15cm}
  \caption{
  Representative image samples in the benchmark dataset.
  }
  \label{fig:eight}
  \vspace{-0.25cm}
\end{figure*}

where $N$ is the batch size, ${{\mathbf{P}}}_B^i$ denotes $i$-th prediction in a mini-batch, and $\mathbf{G}_B^i\in\left\{0,1\right\}^{1\times H\times W}$ is the corresponding true label. Moreover, the BCE loss between the edge output and the corresponding pixel-level label can be formulated as follows:
\begin{equation}
\begin{aligned}
\mathcal{L}_{\mathrm{BCE}}^E    
&=\frac{1}{N} \sum_{i=1}^N \left\{-\frac{1}{HW} \sum_{(h, w)}\left[\mathbf{G}_E^i(h, w) \times \log \left({\mathbf{P}}_E^i(h, w)\right)\right]\right.\\
&-\frac{1}{HW} \left.\sum_{(h, w)}\left[\left(1-\mathbf{G}_E^i(h, w)\right) \times \log \left(1-{\mathbf{P}}_E^i(h, w)\right)\right]\right\},
\end{aligned}
\end{equation}
where $\mathbf{P}_E^i$ denotes $i$-th boundary prediction in a mini-batch, and $\mathbf{G}_E^i\in\left\{0,1\right\}^{1\times H\times W}$ represents the corresponding ground-truth.

On the other hand, the number of crack pixels in photographs is significantly lower than the number of non-crack pixels since most cracks are fine and unnoticeable. Therefore, the crack/non-crack pixels are of a highly imbalanced distribution, and fundamentally, the crack prediction belongs to a class imbalance problem. Dice loss \cite{Dice} concentrates on measuring the overlap and similarity between the prediction and ground-truth. It can thereby alleviate the class-imbalance issue because it is insensitive to the amount of foreground and background pixels \cite{Dice}. Consequently, the Dice loss is also adopted to train networks and can be defined by
\begin{equation}
\mathcal{L}_{\text {Dice }}^B=\frac{1}{N} \sum_{i=1}^N\left\{1-\frac{2 \sum_{(h, w)}\left[{\mathbf{P}}_B^i \times \mathbf{G}_B^i\right]+\epsilon}{\sum_{(h, w)}\left[{\mathbf{P}}_B^i\right]+\sum_{(h, w)}\left[\mathbf{G}_B^i\right]+\epsilon}\right\},
\end{equation}
where $\epsilon$ is a Laplace smoothing item to avoid zero division and is set to ${10}^{-6}$. Furthermore, the identification of crack edges is also a class imbalance problem, so the Dice loss function can also be employed as a metric to represent the overlap between $\mathbf{P}_E^i$ and $\mathbf{G}_E^i$:
\begin{equation}
\mathcal{L}_{\text {Dice }}^E=\frac{1}{N} \sum_{i=1}^N\left\{1-\frac{2 \sum_{(h, w)}\left[{\mathbf{P}}_E^i \times \mathbf{G}_E^i\right]+\epsilon}{\sum_{(h, w)}\left[{\mathbf{P}}_E^i\right]+\sum_{(h, w)}\left[\mathbf{G}_E^i\right]+\epsilon}\right\},
\end{equation}

Further, the gradients of images can represent the pixel changes in images, and the gradients in the crack edge areas are greater than those in the interior portions. In order to improve the representation power of the global output to crack edges and enhance the prediction accuracy, a Charbonnier loss function \cite{8100101} is applied to assess the gradient similarity:
\begin{equation}
\mathcal{L}_{\mathrm{Grad}}=\frac{1}{N} \sum_{i=1}^N \frac{1}{HW} \sum_{(h, w)}\left[\sqrt{\left(\mathbf{P}_G^i-\mathbf{G}_G^i\right)^2+\varepsilon^2}\right],
\end{equation}
where $\mathbf{P}_G^i$ is the $i$-th gradient image and obtained by conducting the Scharr operator on the predicted output ${{\mathbf{P}}}_B^i$, and $\mathbf{G}_G^i$ denotes the $i$-th ground-truth gradient image and achieved by conducting the Scharr operator on the ground-truth label $\mathbf{G}_B^i$ similarly. $\varepsilon$ is set to ${10}^{-3}$ following  \cite{8100101}.

Ultimately, the overall loss $\mathcal{L}_{Total}$ is a mixed loss function combing aforementioned loss functions:
\begin{equation}
\mathcal{L}_{\text {Total }}=\alpha_1 \mathcal{L}_{\mathrm{BCE}}^B+\alpha_2 \mathcal{L}_{\mathrm{BCE}}^E+\alpha_3 \mathcal{L}_{\text {Dice }}^B+\alpha_4 \mathcal{L}_{\text {Dice }}^E+\alpha_5 \mathcal{L}_{\mathrm{Grad}},
\end{equation}
where $\alpha_1$, $\alpha_2$,$ \alpha_3$, $\alpha_4$ and $\alpha_5$ are five weighting hyperparameters to regulate the impact of different sub-loss functions and are set to 1 simply.

%-------------------------------------------------------------------------
\section{Steel Crack Dataset}
The proposed benchmark dataset composed of steel crack images is introduced in this section, which is named Steelcrack for convenience of expression. 

To enhance the practicality of datasets, all the images in these datasets are directly captured from multifarious practical engineering structure projects in diverse scenarios, such as Nanjing Second Yangtze River Bridge, Humen Bridge, etc. Consequently, these images have different backgrounds and crack characteristics regarding length, width, and color. Moreover, defect images are typically captured by workers or inspection equipment in actual engineering, so poor imaging quality is inevitable. In order to adapt to these real-world shooting conditions, images with varying levels of blurring are included in the dataset. Some typical and representative images coming from the dataset are shown in Fig. \ref{fig:eight}. 

After obtaining defect images, flip and rotation tricks are adopted for data augmentation. Then, small patches are cropped from the original images to form the Steelcrack, following the settings of other open-source datasets. Specifically, all the images have a fixed pixel resolution, 512 × 512, and the numbers of images of the training set, validation set, and test set are 3300, 525, and 530, respectively. Details are listed in Table \ref{tab:tab1} above as well.

\begin{table}\footnotesize
    \centering
    \caption{Details of the proposed datasets.}
    \vspace{-0.25cm}
    \resizebox{\linewidth}{!}{
    \begin{tabular}{l|cccc}
      \toprule
      \textbf{\thead{Dataset \\ name}} & \textbf{\thead{Image\\Size}} & \textbf{\thead{Number of images\\
of the training set}} & \textbf{\thead{Number of images\\ of the validation set}} & \textbf{\thead{Number of images\\ of the test set}}\\
      \midrule
      Steelcrack & 512 × 512 & 3300 & 525 & 530\\
      \bottomrule
    \end{tabular}}
    % \vspace{-0.35cm}
    \label{tab:tab1}
    \vspace{-0.55cm}
  \end{table}

%-------------------------------------------------------------------------
\section{Implementation Details}
\subsection{Training and Inference Environment}
The hardware environment of training and testing algorithms is as follows: Intel Xeon E5-2697 @ 2.70 GHz CPU, 3 $\times$ Nvidia GeForce RTX 2080Ti GPUs, and 256 GB RAM. The main software environment: Ubuntu 18.04 operating system, CUDA 11.6.2, CUDNN 8.6.0, and Python 3.8.15. The deep learning framework employs the PyTorch developed by Meta, and the version is 1.12.1.

\subsection{Training Policy}
The Adam optimizer is utilized for training networks, and the hyper-parameters are set to the default values that are $\beta=\ \left(0.9,\ 0.999\right)$, $\varepsilon={10}^{-8}$ and $weight\ decay=0$. The batch size is set as 9, and the learning rate is fixed to $6\times{10}^{-3}$ unless specified otherwise. Networks are all trained for 70 epochs before inference. For simplicity, no additional data augmentation tricks and learning rate schedulers are adopted in the training stage. It is noteworthy that all the models are implemented in this unified training pipeline for equitable and standardized comparisons.

\subsection{Evaluation Metrics}
In order to validate and evaluate the performance of proposed models, two types of metrics are adopted, namely, effect and efficiency metrics.

\textbf{Effect metrics.} In pixel-level segmentation tasks, class-wise metrics, the mean intersection over union (mIoU) and Dice coefficient (Dice), are the two most widely used metrics. However, for the crack detection task, engineers pay more attention to the volume of defects in a single image, so the performance of algorithms on each image is more worthy of attention. Therefore, mIoU and Dice are modified to image-wise metrics and named the mean image-wise Intersection over Union (mi IoU) and mean image-wise Dice coefficient (mi Dice), respectively. The calculation process of them is introduced as follows:

mi IoU. Let the $n$-th predicted results of a model be ${\widetilde{\mathbf{P}}}^n\in\left(0,1\right)^{1\times H\times W},n\in\left[1,N\right]$, here $N$ denotes the number of images to be evaluated. As a binary classification problem, pixels are firstly divided into positive and negative pixels with the threshold segmentation method, and 0.5 is usually used as the classification threshold:
\begin{equation}
\text { Class }_{\widetilde{\mathbf{P}}_{i, j}^n}=\left\{\begin{array}{cl}
0 \text { means Negative } & \tilde{\mathbf{P}}_{i, j}^n \in(0,0.5) \\
1 \text { means Postive } & \widetilde{\mathbf{P}}_{i, j}^n \in[0.5,1)
\end{array},\right.
\end{equation}
where ${\widetilde{\mathbf{P}}}_{i,j}^n\left(i\in\left[1,H\right],j\in\left[1,W\right]\right)$ means the output value in spatial position $\left(i,j\right)$. On the other hand, the ground-truth of the $n$-th image be $\mathbf{G}^n\in\left\{0,1\right\}^{1\times H\times W}$. By comparing ${\rm Class}_{{\widetilde{\mathbf{P}}}_{i,j}^n}$ and $\mathbf{G}^n$,$ {TP}^n$, ${FN}^n$ and ${FP}^n$ can be obtained, where ${TP}^n$ represents the number of true positive pixels, ${FN}^n$ denotes the number of false negative pixels, and ${FP}^n$ is the number of false positive pixels. Then, mi IoU be obtained:
\begin{equation}
\operatorname{mi} \mathrm{IoU}=\frac{1}{N} \sum_{n=1}^N \frac{T P^n+\varepsilon}{F N^n+F P^n+T P^n+\varepsilon},
\end{equation}
here $\varepsilon$ is a constant with a small value to prevent the denominator from being zero and is set to ${10}^{-6}$.

mi Dice. mi Dice be defined by the following formula: 
\begin{equation}
\text { mi Dice }=\frac{1}{N} \sum_{n=1}^N \frac{2\left|\tilde{\mathbf{P}}^n \otimes \mathbf{G}^n\right|+\varepsilon}{\left|\tilde{\mathbf{P}}^n\right|+\left|\mathbf{G}^n\right|+\varepsilon}.
\end{equation}

Obviously, mi IoU and mi Dice are a discrete evaluation metric and a continuous evaluation metric, respectively, and both of them have a positive correlation with the models' performance. Additionally, mi Dice is more interested in the similarity at the image level, whereas mi IoU is inclined to explore the pixel-level performance of predictions. Consequently, combining both of them will result in a more accurate and thorough evaluation of model performance.

\textbf{Efficiency metrics.}
Two measures are presented in this research to assess the models' computational consumption. One is "\#Param." which is used to gauge the size of a deep learning model and the value of \#Param. is equal to the number of parameters in a model. Multiply-accumulate operations (MACs) are another indicator used to assess networks' computational expense. Ptflops, an easy-to-use open-source tool for PyTorch, is the only tool to be used to calculate the values of these two efficiency indexes in this paper to maintain consistency in evaluation criteria.

%-------------------------------------------------------------------------
\section{Experiment}
\subsection{Comparison with Representative Segmentation Networks}
\begin{figure*}[tb!]
  \centering
  \includegraphics[width=0.95\linewidth]{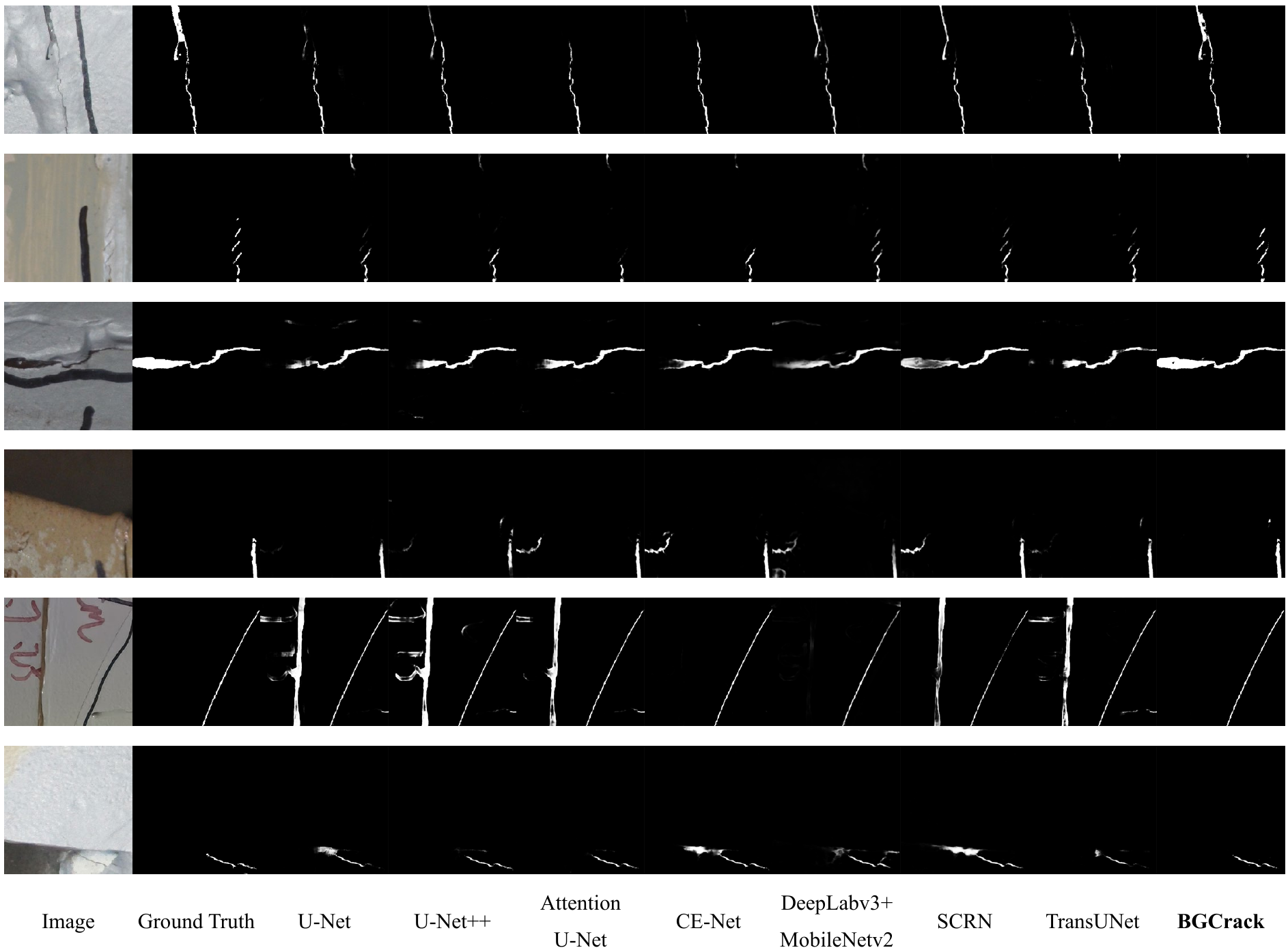}
  \vspace{-0.15cm}
  \caption{
  Visualization results of different models.
  }
  \label{fig:nine}
  \vspace{-0.25cm}
\end{figure*}

BGCrack is compared with some influential segmentation algorithms, such as U-Net and its classical variants, DeepLabv3+ \cite{Deeplab}, SCRN \cite{SCRN}, and TransUNet \cite{chen2021transunet}, from the perspectives of quantitative and qualitative analysis. The corresponding results are listed in Table \ref{tab:tab2} and summarized in Fig. \ref{fig:nine}, respectively.

\textbf{Quantitative comparison.} Table \ref{tab:tab2} shows the detailed comparison results and the highest scores are bolded. It is clear that BGCrack, while using fewer parameters and calculations, is still able to outperform all other models in terms of all assessment measures.

\begin{table}\footnotesize
    \centering
    \caption{Quantitative evaluation of different methods.}
    \vspace{-0.25cm}
    \resizebox{\linewidth}{!}{
    \begin{tabular}{l|cccc}
      \toprule
      \textbf{Method} & \textbf{mi IoU (\%)} & \textbf{mi Dice (\%)} & \textbf{\#Param. (M)} & \textbf{MACs (G)}\\
      \midrule
      U-Net \cite{UNET} & 68.49 & 75.13 & 7.77 & 55.01\\
      U-Net (large)  \cite{UNET} & 69.81 & 76.85 & 31.04 & 219.01\\
      U-Net++ \cite{unetpp} & 72.23 & 78.37 & 9.16 & 138.63\\
      Attention U-Net \cite{attunet} & 71.25 & 77.54 & 34.88 & 266.54\\
      CE-Net\cite{cenet} & 76.00 & 81.54 & 29.00 & 35.60\\
      \vtop{\hbox{\strut DeepLabv3+}\hbox{\strut MobileNetv2\cite{Deeplab}}} & 68.22 & 71.07 & 5.81 & 29.13\\
      \vtop{\hbox{\strut DeepLabv3+}\hbox{\strut xception\cite{Deeplab}}} & 67.40 & 71.48 & 54.70 & 83.14\\
      \vtop{\hbox{\strut DeepLabv3+}\hbox{\strut ResNet-101\cite{Deeplab}}} & 69.04 & 49.45 & 59.34 & 88.84\\
      SCRN \cite{SCRN} & 73.23 & 78.91 & 25.23 & 31.29\\
      TransUNet \cite{chen2021transunet} & 64.34 & 72.55 & 67.87 & 129.96\\
      \midrule
      \textbf{BGCrack} & \textbf{77.16} & \textbf{85.33} & \textbf{2.32} & \textbf{15.76}\\
      \bottomrule
    \end{tabular}}
    % \vspace{-0.35cm}
    \label{tab:tab2}
    \vspace{-0.55cm}
  \end{table}

\textbf{Qualitative visualization results.} As shown in Fig. \ref{fig:nine} below, BGCrack can accurately and consistently identify cracks with different backgrounds, features, and artificial markers compared to other deep-learning models. Besides, the predicted cracks of BGCrack are more transparent and have less noise than that of other methods. The shape of the predicted cracks is also more in line with the ground truth. This directly shows the gain effect of boundary information on recognition results. In general, the visualization results reflect BGCrack's strong robustness and generalization ability.

\subsection{Ablation Study}
Ablation experiments are conducted to validate the rationality of the design.

\textbf{Ablation of the boundary feature modeling.} One of the main contributions of this paper is to introduce crack boundary features to crack identification. An ablation experiment is conducted to explore the effectiveness of this idea. Specifically, parts and modules related to the edge feature modeling are removed from the BGCrack, and the remaining parts are composed of a new sub-network. Then, it is trained using the total loss function removing $\mathcal{L}_{BCE}^E$ and $\mathcal{L}_{Dice}^E$. The corresponding evaluation results are shown in Table \ref{tab:tab3}. Obviously, considering edge features is helpful to segment cracks accurately based on the comparison in Table \ref{tab:tab3}. We hope this study can inspire researchers to conduct broader and deeper research in this direction.

\begin{table}\footnotesize
    \centering
    \caption{Ablation studies of the boundary feature modeling.}
    \vspace{-0.25cm}
    \resizebox{\linewidth}{!}{
    \begin{tabular}{l|cccc}
      \toprule
      \textbf{Edge} & \textbf{mi IoU (\%)} & \textbf{mi Dice (\%)} & \textbf{\#Param. (M)} & \textbf{MACs (G)}\\
      \midrule
      \usym{1F5F4} & 73.57 & 82.57 & 1.94 & 8.17\\
      \usym{1F5F8} & 77.16 & 85.33 & 2.32 & 15.76\\
      \bottomrule
    \end{tabular}}
    % \vspace{-0.35cm}
    \label{tab:tab3}
    \vspace{-0.3cm}
  \end{table}

\textbf{Effectiveness of the gradient loss function.} The effectiveness of the gradient loss function $\mathcal{L}_{Grad}$ is evaluated here. We take the other 4 loss functions keeping the same weighting hyperparameters to train the BGCrack and record the corresponding results in Table \ref{tab:tab4} above. Based on the data in Table \ref{tab:tab4}, the performance of the model is reduced if the gradient loss function is removed from the total loss, which verifies the availability of the gradient loss.

\begin{table}\footnotesize
    \centering
    \caption{Effectiveness validation of the gradient loss function.}
    \vspace{-0.25cm}
    \resizebox{\linewidth}{!}{
    \begin{tabular}{l|cccc}
      \toprule
      \textbf{Loss} & \textbf{mi IoU (\%)} & \textbf{mi Dice (\%)} & \textbf{\#Param. (M)} & \textbf{MACs (G)}\\
      \midrule
      \usym{1F5F4} & 76.53 & 84.85 & 2.32 & 15.76\\
      \usym{1F5F8} & 77.16 & 85.33 & 2.32 & 15.76\\
      \bottomrule
    \end{tabular}}
    % \vspace{-0.35cm}
    \label{tab:tab4}
    \vspace{-0.3cm}
  \end{table}

\textbf{Ablation experiments of the HFIE and GIP modules.} 	The HFIE and GIP modules are adopted to enhance high-frequency information and global perception, respectively. They are both lightweight and plug-and-play modules. In this subsection, ablation experiments are conducted to validate the effectiveness of the two modules. As shown in Table \ref{tab:tab5}, HFIE and GIP can both enrich representation ability with negligible computational overheads.

\begin{table}\footnotesize
    \centering
    \caption{Effectiveness of the HFIE and GIP.}
    \vspace{-0.25cm}
    \resizebox{\linewidth}{!}{
    \begin{tabular}{l|cccc}
      \toprule
      \textbf{Model} & \textbf{mi IoU (\%)} & \textbf{mi Dice (\%)} & \textbf{\#Param. (M)} & \textbf{MACs (G)}\\
      \midrule
      -HFIE-GIP & 74.06 & 82.93 & 2.16 & 15.54\\
      -GIP & 74.37 & 83.27 & 2.19 & 15.68\\
      -HFIE & 75.89 & 84.21 & 2.28 & 15.62\\
      BGCrack & 77.16 & 85.33 & 2.32 & 15.76\\
      \bottomrule
    \end{tabular}}
    % \vspace{-0.35cm}
    \label{tab:tab5}
    \vspace{-0.3cm}
\end{table}

\subsection{Visualization Analysis}
\begin{figure*}[tb!]
  \centering
  \includegraphics[width=0.95\linewidth]{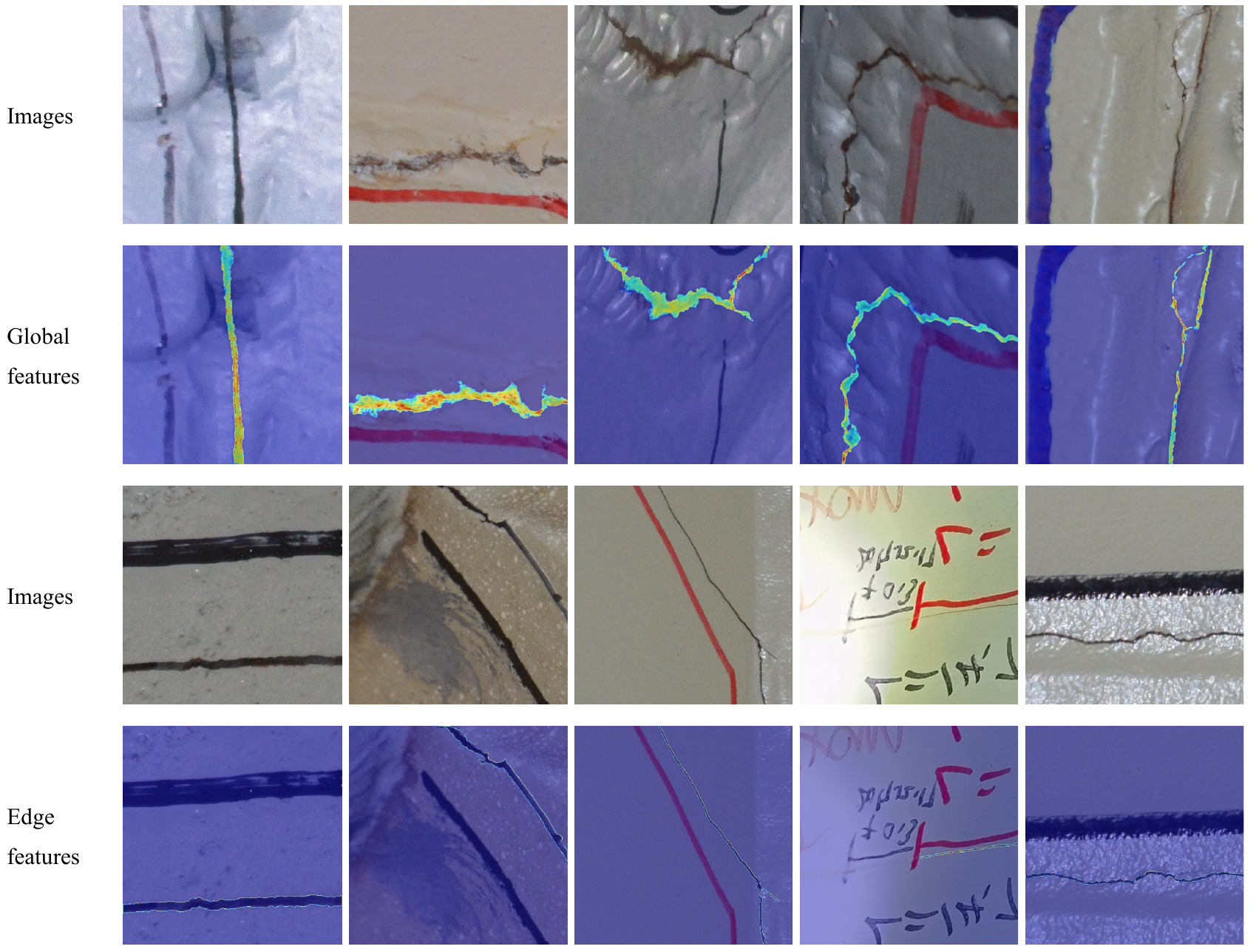}
  \vspace{-0.15cm}
  \caption{
  Visualization feature maps generated by Grad-CAM.
  }
  \label{fig:ten}
  \vspace{-0.25cm}
\end{figure*}

To delve into if BGCrack can pay attention to defects areas correctly, the feature maps are visualized by Grad-CAM \cite{Gradcam}, which is one of the most influential visual explanation tools and can reveal key evidence and reasons that models use for reference. The heat maps of global and boundary features are presented in Fig. \ref{fig:ten}. Obviously, BGCrack focuses on the crack locations correctly, which provides a visual explanation for the high recognition accuracy of the model.

% \vspace{-0.25cm}
%------------------------------------------------------------------------
\section{Conclusion}
% \vspace{-0.15cm}

Cracks are one of the critical signs of structural performance degradation, and improving the accuracy of crack identification is an issue that has attracted extensive attention. The conventional processing method uses advanced deep learning models directly, such as Transformer, but neglects the inherent characteristics of this problem. Therefore, this paper tries to combine the characteristics with deep learning techniques and design a targeted algorithm named BGCrack. The main contributions of this paper are as follows:

(1) BGCrack decouples cracks into two parts: the crack boundary and internal area, and integrates boundary information into segmentation to generate more accurate and clear predictions. Targeted modules, such as HFIE, GIP, and COM, are designed to enhance the model’s representation power. Extensive experiments have been conducted to validate the effectiveness of the proposed design.

(2) Because steel structures are one of the main structural forms in civil engineering, fatigue cracks in steel structures are very common during service and are important safety hazards. This paper provides a steel crack dataset to promote the research on the identification of steel cracks.

The current pipeline still has several limitations; hence it is suggested that further research might be conducted in the following directions: (1) Improving the performance of inspection algorithms further. Although BGCrack outperforms other advanced segmentation frameworks, there is still room for improvement with respect to inspection accuracy. The following research plan is to develop more robust detectors with better generalization capability using cutting-edge AI technologies. (2) The existing framework focuses on intelligent pixel-level segmentation, which is just the first step of defect detection. Defect quantification tools will be integrated into this framework to build a more complete and thorough inspection procedure. (3) In recent days, large language models (LLM) represented by Chat-GPT and GPT-4 have drawn extensive attention from all over the world and pushed the research of AI to a peak. It is a promising direction for developing Civil-GPT, integrating intelligent planning, structural design, construction, and maintenance to assist civil engineers in making scientific decisions and reduce engineers’ time overhead.

\textbf{Declaration of competing interest.} The authors declare that they have no known competing financial interests or personal relationships that could have appeared to influence the work reported in this paper.
\newline

\textbf{Acknowledgements.} The research presented was financially supported by the National Key R\&D Program of China [grant number 2022YFC3801700], the Innovation Technology Fund, Midstream Research Programme for Universities [grant number MRP/003/21X], and the Hong Kong Research Grants Council [grant number 16205021]. The authors would like to thank Prof. Yang Xu and Prof. Hui Li of Harbin Institute of Technology for providing part of the crack images. Finally, contributions by the anonymous reviewers are also highly appreciated.
\newline

% \vspace{0.2cm}
% \noindent \textbf{Acknowledgement. }
% The work is supported by the NSFC Program (62222604, 62206052, 62192783), CAAI-Huawei MindSpore (CAAIXSJLJJ-2021-042A), China Postdoctoral Science Foundation Project
% (2021M690609), Jiangsu Natural Science Foundation
% Project (BK20210224), and CCF-Lenovo Bule Ocean
% Research Fund.

\clearpage

%%%%%%%%% REFERENCES
{\small
\bibliographystyle{ieee_fullname}
\bibliography{egbib}
}

\end{document}